\documentclass{article}

\usepackage{iclr2027_conference,times}

\usepackage{amsmath,amsfonts,bm}

\def\eqref#1{equation~\ref{#1}}

\def\1{\bm{1}}

\DeclareMathAlphabet{\mathsfit}{\encodingdefault}{\sfdefault}{m}{sl}
\SetMathAlphabet{\mathsfit}{bold}{\encodingdefault}{\sfdefault}{bx}{n}

\usepackage{hyperref}
\usepackage{url}
\usepackage{graphicx}
\usepackage{wrapfig}
\usepackage{amssymb}
\usepackage{amsmath}
\usepackage{colortbl}
\usepackage{xcolor}
\usepackage{newfloat}
\usepackage{listings}
\usepackage{amsthm}

\usepackage{booktabs}
\usepackage{multirow,makecell}
\usepackage{array}
\usepackage{tabularx}
\usepackage{algorithm}
\usepackage{algpseudocode}

\title{Not Every Change Is Necessary: Recoverable Drift in Large Language Model Unlearning}

\author{
\begin{tabular}{@{}c@{}}
Xunlei Chen\textsuperscript{1},
Qinghui Gong\textsuperscript{2},
Jingkun Xue\textsuperscript{1},
Qihe Liu\textsuperscript{1},
Shijie Zhou\textsuperscript{1},
Fei Ye\textsuperscript{1,*}\\
{\normalfont\small
\textsuperscript{1}University of Electronic Science and Technology of China},
{\normalfont\small
\textsuperscript{2}Southwest Jiaotong University}\\
{\normalfont\small
\textsuperscript{*}Corresponding author:
{\texttt{feiye@uestc.edu.cn}}}
\end{tabular}
}

\iclrfinalcopy

\usepackage{etoolbox}
\makeatletter
\patchcmd{\@maketitle}
  {\hsize\textwidth}
  {\hsize\textwidth\centering}
  {}
  {\PackageWarning{arxiv-layout}{Could not center the title block}}
\makeatother

\fancypagestyle{arxivfirst}{
    \fancyhf{}
    \fancyfoot[L]{\normalfont\footnotesize Under review.}
    \fancyfoot[C]{\thepage}
    \renewcommand{\headrulewidth}{0pt}
    \renewcommand{\footrulewidth}{0pt}
}

\begin{document}

\maketitle

\pagestyle{fancy}
\fancyhf{}
\fancyfoot[C]{\thepage}
\renewcommand{\headrulewidth}{0pt}
\renewcommand{\footrulewidth}{0pt}

\thispagestyle{arxivfirst}





\begin{abstract}
Machine unlearning in large language models aims to remove unwanted knowledge while preserving the model's remaining capabilities. Although existing methods use retention objectives or restrict where edits occur, achieving the desired forgetting level can still leave collateral changes that impair non-target behavior. Our recovery comparisons suggest that some of these changes can be reversed while preserving observed forgetting performance. 
In this work, we present \textbf{\underline{P}}ropose-\textbf{\underline{T}}hen-\textbf{\underline{P}}roject Unlearning (PTP-U), a framework that combines targeted forgetting with the recovery of non-target capabilities.
PTP-U first applies local analytic edits to weaken target knowledge associations, then aligns non-target output distributions with those of the original model to recover capabilities while maintaining fixed forgetting constraints. Both stages serve a common goal: satisfying the forgetting requirements while preserving fluent generation and performance on non-target tasks. Across three benchmarks, PTP-U achieves the strongest forgetting–retention trade-off among evaluated methods, reaching 81.22\%–91.03\% forgetting while preserving 94.20\% non-target utility on average. At matched forgetting, PTP-U consistently retains higher non-target utility. 

\end{abstract}

\section{Introduction}

Large language models (LLMs) acquire extensive knowledge, skills, and linguistic patterns during pre-training~\citep{pan2026understanding, guo-etal-2025-hash, ranjan2026razor}, becoming general-purpose models for long-term deployment~\citep{zhao2025qwen3guard, ye2026learning1}. However, training corpora may include private information~\citep{ramakrishna2025lume}, copyrighted content~\citep{shi2025muse}, or unsafe knowledge~\citep{li2024wmdp}, requiring the removal of specific information after training~\citep{yang2026erase}. While retraining from scratch after excluding target data can achieve this goal, it is computationally infeasible for modern LLMs~\citep{gong2026dssdynamicsemanticsteering,ye2026learning2}. Therefore, LLM unlearning aims to selectively remove target knowledge while preserving the model's original capabilities and behaviors on non-target inputs.

Existing unlearning methods mainly apply forgetting signals at the output level by suppressing target content generation~\citep{pu2026decoding}, adjusting output distributions~\citep{ma2026retrace}, or optimizing preference objectives~\citep{fan2025simplicity}. They are often combined with retain objectives or regularization to preserve non-target behaviors.
However, output-level objectives offer limited control over which parameters change and by how much, making it difficult to confine updates to parameters associated with target knowledge~\citep{wang2025model}.
Another line of methods directly modifies internal representations or local parameters to weaken target knowledge associations. These methods control modification scopes through low-rank updates~\citep{chen2026alter}, local constraints~\citep{shen2026llm}, or analytical editing~\citep{li2026cross}. Although internal intervention more directly affects computations related to target knowledge, local changes still propagate nonlinearly through deep Transformer architectures, making global behavioral changes difficult to fully control.

\begin{wrapfigure}{r}{0.52\textwidth}
\centering
\vspace{-10pt} 
\includegraphics[width=0.99\linewidth]{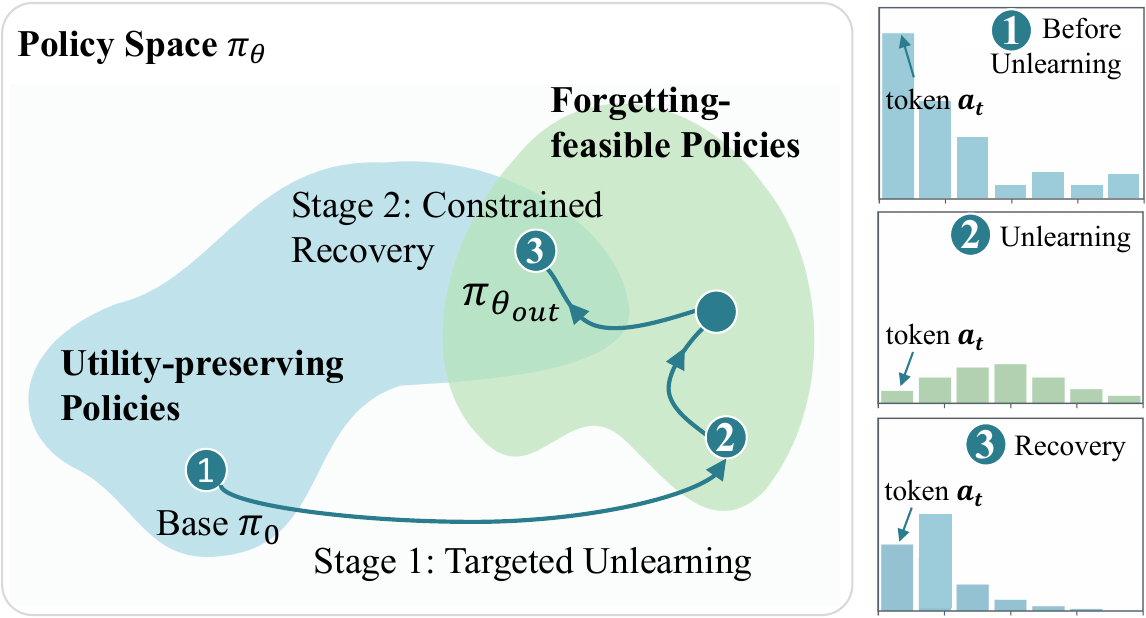}
\caption{PTP-U first achieves forgetting feasibility (Stage I), then restores utility while maintaining feasibility (Stage II). Right: target-answer suppression persists during recovery.}
\label{Fig_Intro}
\vspace{-10pt} 
\end{wrapfigure}

Therefore, achieving the desired forgetting level does not imply that all accompanying changes are necessary. Some changes may be redundant perturbations from optimization, disrupting the model's semantic structure and causing semantic drift or capability degradation~\citep{pmlr-v267-wang25en,lee2026continual}. \textbf{This motivates a key question: can such collateral changes be reversed while preserving forgetting?}
To explore this, we investigate recoverable changes in models that have achieved the target forgetting level. We adjust the unlearned model toward the Base model while retaining the forgetting effect. As shown in Figure~\ref{Fig_Intro}, if certain changes can be restored without recovering the forgotten knowledge, they are unnecessary for unlearning and instead reflect optimization-induced drift. We define such reversible changes as \textit{recoverable drift}.

Prior post-training studies show that among models solving the same new task, solutions with output behaviors closer to the Base model retain more original capabilities~\citep{shenfeld2026rl}. Inspired by this foundation, we propose Propose-Then-Project Unlearning (PTP-U), which combines local analytical editing with constrained recovery fine-tuning. 
PTP-U first constructs candidate models through local analytic edits and then reduces behavioral drift by aligning non-target outputs with the Base model under fixed forgetting constraints.

Specifically, PTP-U addresses the same constrained objective at two scales. In the local proposal stage, we use Fisher geometry to approximate Base-policy drift quadratically and efficiently generate candidate updates under linearized forgetting constraints. 
When updates move beyond the local region, Transformer nonlinearities weaken this approximation. Thus, the nonlinear policy projection stage operates directly on output distributions to reduce deviation from the Base policy under fixed forgetting constraints.
For sequential Transformer modules, PTP-U re-estimates the local geometry and constraints of downstream components after upstream updates, reducing linearization errors. This separates local proposal generation from subsequent nonlinear recovery while retaining a common objective and fixed forgetting requirements.
Across three benchmarks, PTP-U achieves the strongest forgetting/retention trade-off, reaching 81.22\%–91.03\% forgetting (compared with 80.70\%--87.93\% for the strongest forgetting baseline) while preserving the highest average capability retention of 94.20\%. At matched forgetting, it consistently retains higher utility (93.32\%–95.86\%) and remains closer to the Base model. These results support our central finding that a substantial portion of the changes induced by strong unlearning is not necessary for maintaining forgetting. In summary, our contributions are: 

(1) We propose Propose-Then-Project Unlearning (PTP-U), a Base-centered framework that couples targeted forgetting with explicit behavioral recovery. It separates forgetting-oriented proposal generation from recovery toward the Base policy under fixed forgetting constraints.

(2) We develop an analytic editing mechanism that constructs local proposals under linearized forgetting constraints. Sequential relinearization updates downstream geometry and constraints after upstream edits, accounting for dependencies within Transformer blocks.

(3) We validate PTP-U across multiple benchmarks and model families, demonstrating strong forgetting--retention trade-offs, robustness against prompt-based recovery and relearning attacks, and stable performance across diverse settings. 

\vspace{-10pt}
\section{Recoverable Drift: From Observation to Formulation}
\vspace{-5pt}

\subsection{Forgetting Constraints and Behavioral Drift}

Let \(\pi_0=\pi_{\theta_0}\) be the Base model, \(\pi_\theta\) the updated model, and \(\mathcal U_f\) the set of target knowledge units. For each \(u\in\mathcal U_f\), we measure relation support using a target query \(q\), a matched contrastive query \(\bar q\) without the target relation, and an answer \(a\), where \(|a|\) denotes the number of tokens in \(a\):
\begin{equation}
S_u(\theta)
=
\mathbb E_{(q,\bar q,a)\sim\mathcal V_u}
\left[\ell_\theta(a\mid q)-\ell_\theta(a\mid\bar q)\right],
\qquad
\ell_\theta(a\mid q)
=
\frac{1}{|a|}\sum\nolimits_{t=1}^{|a|}
\log\pi_\theta(a_t\mid q,a_{<t}),
\label{Eq_1}
\end{equation}
where \(\mathcal V_u\) contains query views and answer aliases. Lower scores \(S_u(\theta)\) indicate weaker relation support. We fix thresholds \(b_u\) before updating and define the forgetting-feasible set \(\mathcal C_f\), which contains models satisfying all target-unit thresholds:
\begin{equation}
\mathcal C_f
=
\left\{
\theta\mid c_u(\theta)\le0,\ 
\forall u\in\mathcal U_f
\right\},
\qquad
c_u(\theta)=S_u(\theta)-b_u.
\label{Eq_2}
\end{equation}
Sample construction and threshold calibration are detailed in Appendices
On a fixed non-target context distribution $\mu_a$, we define Base-policy drift as
\begin{equation}
D_{\mathrm B}(\theta)
=\mathbb E_{s\sim\mu_a}
D_{\mathrm{KL}}\left(
\pi_0(\cdot\mid s)\,\middle\|\,\pi_\theta(\cdot\mid s)
\right).
\label{Eq_3}
\end{equation}
Here, \(D_{\mathrm{KL}}\) denotes the Kullback--Leibler (KL) divergence, and \(s\) includes the input and its generation prefix. Smaller \(D_{\mathrm B}(\theta)\) indicates closer next-token distributions on these contexts.

\vspace{-5pt}
\subsection{Recoverable Drift and Constrained Unlearning Formulation}
\vspace{-5pt}

\begin{figure}[t]
\centering
\includegraphics[width=0.99\linewidth]{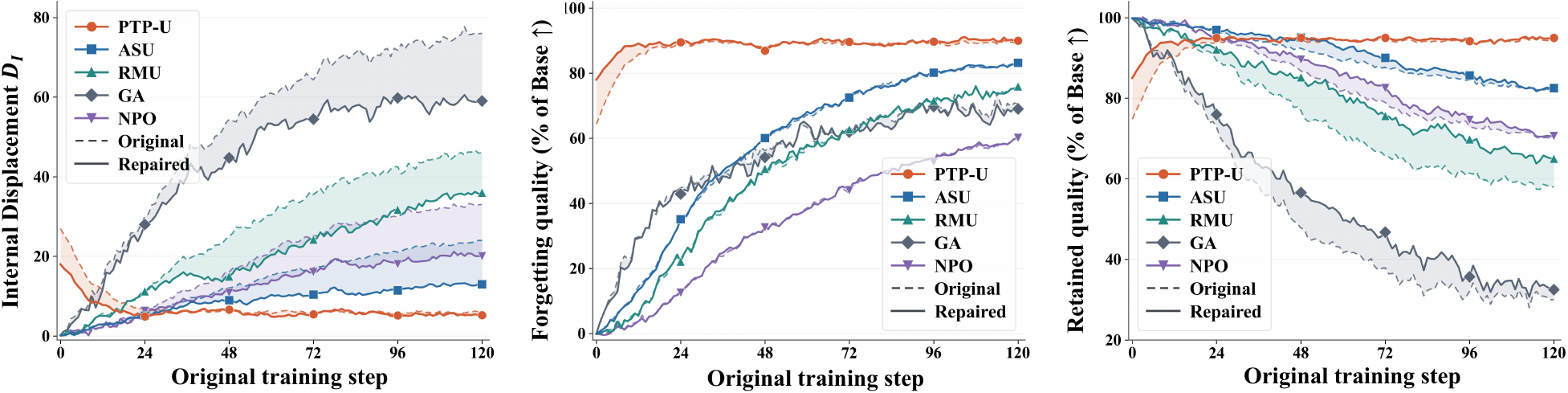}
\vspace{-5pt}
\caption{Checkpoint-wise recovery across unlearning methods. Dashed lines show original checkpoints and solid lines their independently repaired copies. Shading marks paired differences. The horizontal axis records fine-tuning steps. PTP-U step 0 follows the \textit{Stage I} analytic edits. Repairs are performed separately and do not feed into subsequent original checkpoints.}
\label{Fig_Main:Obs}
\vspace{-10pt}
\end{figure}

We examine whether changes introduced during unlearning remain necessary once forgetting has been achieved. Results from Figure~\ref{Fig_Main:Obs} address this question through paired recovery comparisons. To track internal changes, let $h_\theta^\ell(s)$ denote the hidden representation at layer $\ell$ and $P_{\mathrm{ret}}$ the orthogonal projector onto a fixed retain-sensitive subspace. We focus on displacement within this subspace and normalize by the Base representation energy to express changes relative to the original representation scale. We measure internal displacement as
\begin{equation}
D_{\mathrm I}(\theta)
=100\sqrt{
\frac{
\mathbb E_{s\sim\mu_a}
\left\|P_{\mathrm{ret}}\left(h_\theta^\ell(s)-h_{\theta_0}^\ell(s)\right)\right\|_2^2
}{
\mathbb E_{s\sim\mu_a}
\left\|P_{\mathrm{ret}}h_{\theta_0}^\ell(s)\right\|_2^2+\varepsilon
}},
\label{Eq_InternalDisplacement}
\end{equation}
where $\|\cdot\|_2^2$ denotes the squared L2 norm and $\varepsilon>0$ is a fixed numerical stabilizer. The contexts, layer, and projector remain fixed across comparisons. Appendix
specifies subspace construction, metric definitions, and repair settings.

Recovery generally reduces internal displacement while leaving forgetting quality largely unchanged, with retention gains varying across methods. These empirical results support our \textit{recoverable drift} premise: some internal displacement introduced by unlearning can be reduced without undoing forgetting. Accordingly, model selection should consider both forgetting feasibility and behavioral deviation from the Base model. Inspired by this, we treat recovery as an explicit part of the unlearning pipeline, instead of treating all unlearning-induced modifications as indispensable.

While $D_{\mathrm I}(\theta)$ is a representation diagnostic, we use $D_{\mathrm B}$ to measure behavioral deviation directly in output space and formulate our constrained unlearning objective as
\begin{equation}
\underset{\theta \in \Theta}{\operatorname{minimize}}\;
D_{\mathrm B}(\theta)
\qquad \text{s.t.}\quad
S_u(\theta)\le b_u,
\qquad \forall u\in\mathcal U_f,
\label{Eq_5}
\end{equation}
where \(\Theta\) denotes the model's parameter space. This objective motivates an explicit recovery phase while keeping the forgetting thresholds fixed throughout optimization. Our method first applies local analytic edits to seek feasible solutions, then performs constrained recovery to reduce undesirable behavioral drift of the model. Global optimality is not assumed. 

\vspace{-5pt}
\section{Propose-Then-Project Unlearning (PTP-U)}
\vspace{-5pt}

PTP-U approximately solves Eq.~\ref{Eq_5} via two stages. It first generates a forgetting proposal from the local approximation of the current model and verifies a feasible initialization with the actual model. Then, nonlinear policy projection reduces Base-policy drift while maintaining forgetting constraints. Attention$\rightarrow$FFN sequential re-linearization improves the accuracy of local updates in Transformers.

\begin{figure}[t]
\centering
\includegraphics[width=0.99\linewidth]{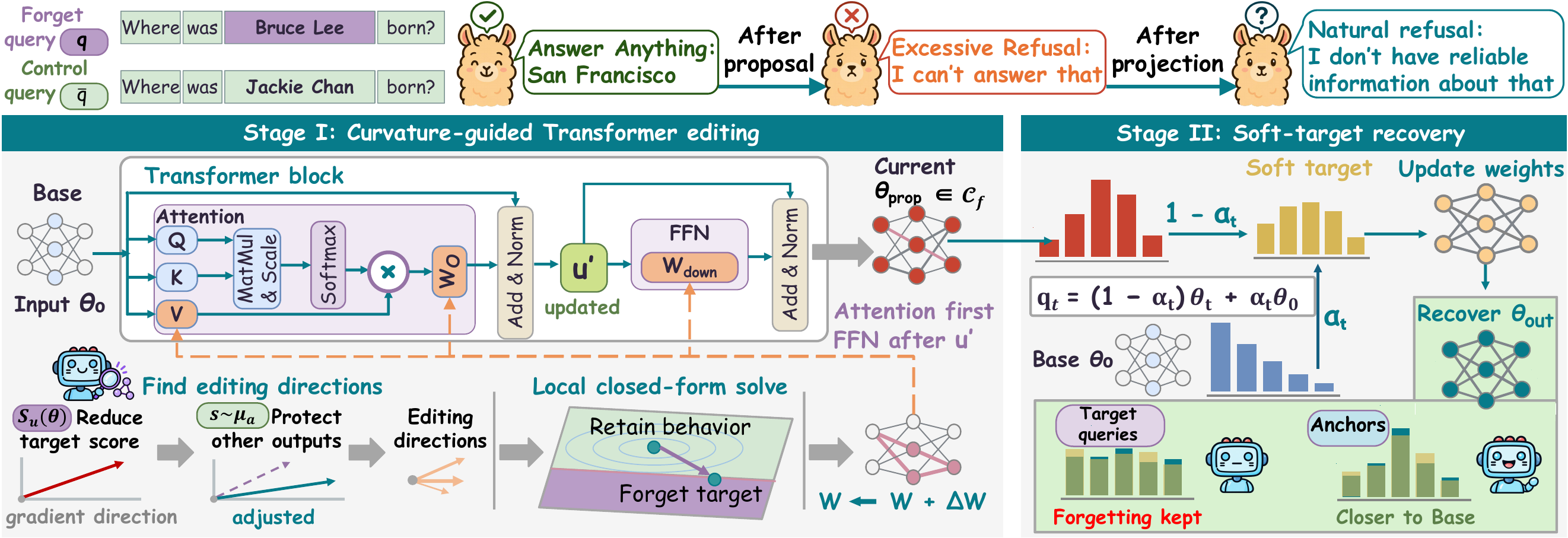}
\vspace{-10pt}
\caption{Overview of PTP-U. \textit{Stage I} computes a curvature-guided closed-form proposal in an editable subspace. It updates Attention first, refreshes the operating point, then updates FFN to obtain a feasible unlearned model. \textit{Stage II} constructs a soft target from current and Base policies for constrained recovery. It preserves forgetting while reducing behavioral drift and excessive refusal.}
\label{Fig_Main:Pipline}
\vspace{-10pt}
\end{figure}

\subsection{Local Analytic Proposal with Curvature Guidance}
Starting from the Base parameters, we construct the update around the current model \(\theta_k\). To reduce computational cost, we restrict the parameter change to \(\delta=U_kz\), where \(U_k\) is a full column rank low-dimensional editable basis and \(z\) is the optimized coefficient vector. The editing directions are derived from target knowledge gradients and preconditioned by the K-FAC~\citep{martens2015optimizing} approximation of the current model Fisher information~\citep{amari1998natural} on fixed anchors, capturing the local influence of different directions on non-target outputs. The same curvature approximation defines the local quadratic model below.
Let \(c_k=c(\theta_k)\) denote the stacked constraint residuals and \(J_k\) the Jacobian with respect to \(z\). We define the local model as
\begin{equation}
m_k(z)=\widetilde g_k^\top z+\frac12z^\top H_kz,
\qquad
c(\theta_k+U_kz)\approx c_k+J_kz,
\label{Eq_6}
\end{equation}
where \(m_k\) approximates the change of \(D_{\mathrm B}\), \(\widetilde g_k=U_k^\top\nabla_\theta D_{\mathrm B}(\theta_k)\), and \(H_k=U_k^\top(\widehat F_{a,k}+\lambda I)U_k\) with \(\lambda>0\). Here, \(F_{a,k}\) is the Fisher information of the current model on fixed anchors, and \(\widehat F_{a,k}\) is its positive semidefinite K-FAC approximation. For the next-token softmax distribution, \(F_{a,k}\) corresponds to the generalized Gauss-Newton curvature with logits as output variables. The damping term improves stability and, together with the full column rank of \(U_k\), ensures \(H_k\succ0\).

The Base gradient vanishes, but $\widetilde g_k$ is generally nonzero at later iterates and is retained in $m_k$.
Appendix
derives the gradient and separates Fisher curvature from the additional logit-curvature term.
Appendix
quantifies the error of the damped surrogate.
To cap the required residual reduction in each iteration, we define the truncated constraint residual as
\begin{equation}
r_{u,k}=\min\{c_u(\theta_k),r_{\max}\},
\qquad
r_{\max}>0.
\label{Eq_7}
\end{equation}
Here, $r_{\max}$ is a hyperparameter that caps the required reduction for unsatisfied constraints. For satisfied constraints, the negative residual is preserved as the boundary margin for linearized feasibility checking. We optimize the coefficients $z$ by minimizing the local objective $m_k(z)$:
\begin{equation}
\underset{z}{\operatorname{minimize}}
\quad m_k(z)
\qquad
\operatorname{subject\ to}
\quad J_kz\le-r_k.
\label{Eq_8}
\end{equation}
Since large positive residuals are truncated, a single iteration may only achieve partial forgetting.
For a fixed active set $A$, let $J_A$ and $r_A$ denote the corresponding constraint matrix and residual. When $J_A$ has full row rank, the unconstrained recovery direction $v_k$ and constraint multipliers $\nu_A$ yield the analytical solution
\begin{equation}
v_k=-H_k^{-1}\widetilde g_k,
\quad
\nu_A=
\left(J_AH_k^{-1}J_A^\top\right)^{-1}
\left(J_Av_k+r_A\right),
\quad
z_k^{\mathrm{prop}}
=v_k-H_k^{-1}J_A^\top\nu_A.
\label{Eq_9}
\end{equation}
Here, $(\cdot)^{-1}$ denotes the matrix inverse. 
If $\nu_A\ge0$ and all remaining inequalities hold, Eq.~\ref{Eq_9} is the unique solution to Eq.~\ref{Eq_8} by Appendix
The first term minimizes the unconstrained local model, and the second gives the minimum $H_k$-metric correction enforcing the active constraints.
Since Eq.~\ref{Eq_9} only solves the local problem, the local model must be reconstructed after the update. This local solution does not provide a global closed-form solution to the nonlinear unlearning problem.

\subsection{Sequential Relinearization in Transformers}

The local proposal considers the computation order of Transformer blocks, formulated as
\begin{equation}
u_l=h_l+A_l(h_l),
\qquad h_{l+1}=u_l+F_l(u_l),
\label{Eq_10}
\end{equation}
where \(A_l(\cdot)\) and \(F_l(\cdot)\) represent the Attention and FFN submodules, respectively. For a fixed block input, updating Attention parameters \(\theta_A\) affects the FFN input through the following relation, where \(J_{F_l}(u_l)\) denotes the FFN input Jacobian. The first-order joint model captures the propagation of Attention changes through FFN.

\begin{equation}
\frac{\partial h_{l+1}}{\partial\theta_A}
=
\left(I+J_{F_l}(u_l)\right)
\frac{\partial A_l(h_l)}{\partial\theta_A}.
\label{Eq_11}
\end{equation}

Therefore, we first solve \(\delta_A\) in the current Attention editing subspace and update parameters as \(\theta_k^A=\theta_k+\delta_A\). Then, we perform a forward pass to refresh the knowledge unit residuals and Base recovery gradients under \(\theta_k^A\), recompute FFN K-FAC curvature statistics on fixed anchors, and update the FFN editing basis with its constraint Jacobian to solve for \(\delta_F\). Relinearization at $\theta_k^A$ removes the stale-Jacobian term from the FFN error decomposition, as shown in Appendix

Each candidate update is evaluated on the actual model to measure forgetting progress and constraint preservation. If the predicted change differs significantly from the actual change or violates previous constraints, the update is rejected, the target advancement magnitude is reduced, and optimization is repeated. 
This process continues until a feasible checkpoint \(\theta_{\mathrm{prop}}\in\mathcal C_f\) is obtained, which serves as the initialization for the next stage.

\subsection{Nonlinear Policy Projection}
The finite-dimensional editing subspace and local curvature approximation constrain proposal accuracy, while larger updates may cause behavioral changes beyond the local model. Thus, the second stage starts from $\theta_{\mathrm{prop}}$ and continues optimizing Eq.~\ref{Eq_5} on the true output distribution. 
At projection iteration t, let \(\theta_t\in\mathcal C_f\) denote the current feasible model. We construct a progressive recovery target towards the Base model on fixed anchors:
\begin{equation}
q_t(\cdot\mid s)
=(1-\alpha_t)\pi_{\theta_t}(\cdot\mid s)
+\alpha_t\pi_0(\cdot\mid s),
\qquad 0<\alpha_t\le1.
\label{Eq_12}
\end{equation}
Here, \(q_t\) is fixed and gradients are stopped during the inner-loop optimization, while \(\alpha_t\) determines the shift of the target distribution toward the Base model. At each round, we minimize \(\mathcal L_t(\theta)\) over the forgetting feasible region \(\theta\in\mathcal C_f\), where $\mathcal L_t(\theta)$ is defined as
\begin{equation}
\begin{aligned}
\mathcal L_t(\theta)
&=\mathbb E_{s\sim\mu_a}
D_{\mathrm{KL}}\!\left(
q_t(\cdot\mid s)\,\middle\|\,\pi_\theta(\cdot\mid s)
\right)\\
&=\alpha_tD_{\mathrm B}(\theta)
+(1-\alpha_t)\mathbb E_{s\sim\mu_a}
D_{\mathrm{KL}}\!\left(
\pi_{\theta_t}(\cdot\mid s)\,\middle\|\,\pi_\theta(\cdot\mid s)
\right)+C_t,
\end{aligned}
\label{Eq_13}
\end{equation}
where $C_t$ is independent of $\theta$.
Minimizing $\mathcal L_t$ over $\mathcal C_f$ defines a proximal step for Eq.~\ref{Eq_5}, combining Base recovery with a penalty on policy changes.
Appendix
establish the decomposition and sufficient descent conditions, respectively.
The optimization is conducted by adaptively updating the Lagrangian multiplier to satisfy the unlearning constraint~\citep{dhillon2025l3ms}, and candidate solutions are verified on the real model. Only updates meeting the following conditions are accepted:
\begin{equation}
\theta_{t+1}\in\mathcal C_f, \qquad D_{\mathrm B}(\theta_{t+1})<D_{\mathrm B}(\theta_t).
\label{Eq_14}
\end{equation}
If verification fails, the update is reverted and the recovery magnitude is reduced. Base only provides the recovery distribution over non-target anchors. Since all inputs share the same model parameters, recovery must preserve the fixed forgetting constraint to avoid restoring target knowledge. When the computational budget is exhausted or no candidate satisfying Eq.~\ref{Eq_14} is found within the tolerance, the accepted feasible model \(\theta_{\mathrm{out}}\) is returned. Here, projection denotes the constrained recovery process in Eq.~\ref{Eq_5}, which does not ensure the globally nearest point. The complete update, rollback, and stopping rules are described in Appendix
The editing basis, curvature statistics, and recovery objective are used only during training and introduce no additional inference modules.

\section{Experiments}
\vspace{-5pt}
\subsection{Experimental Settings}
\vspace{-5pt}
\textbf{Datasets.}
We evaluate PTP-U on three benchmarks.
RWKU evaluates the removal of real-world entity knowledge~\citep{cao2024rwku}.
WMDP measures hazardous knowledge in biosecurity (Bio) and cybersecurity (Cyber) ~\citep{li2024wmdp}.
MUSE-Books evaluates copyright unlearning on the Harry Potter corpus~\citep{shi2025muse}.
MMLU measures the preservation of general knowledge~\citep{hendrycks2020measuring}.
Dataset and split details are detailed in Appendix

\textbf{Evaluation Metrics.}
For RWKU, we report ROUGE-L recall on fill-in-the-blank (FB), question-answering (QA) probes, and adversarial attack (AA) scenarios. For WMDP, we report accuracy on Bio/Cyber. The 25\% random-choice indicates best forgetting. For MUSE-Books, we report BLEU and ROUGE-L. 
MMLU accuracy (Gen.) and GPT-4o fluency~\citep{xu2025obliviate} assess retained utility. 
We report R-Ent. separately on RWKU and WMDP to measure confirmed restricted knowledge re-Entry after an initial risk closure. Metric details are in Appendix

\textbf{Baselines.}
We compare PTP-U with NPO\_KL~\citep{zhang2024negative}, RMU~\citep{li2024wmdp}, 
WHP~\citep{eldan2023s},
ALTER~\citep{chen2026alter},
Attention-Smoothing (ASU)~\citep{zare2026attention} and ICUL~\citep{pawelczyk2024context},
adapted SCANS~\citep{cao2025scans}, MET~\citep{yu2025unierase}. 
The details of Baselines are provided in Appendices

\textbf{Implementation Details.}
Across all backbones, we set \(r_{\max}=0.25\), \(\lambda=1.0\times10^{-3}\), and \(\alpha_{t}=0.5\) as the initial recovery mixing coefficient. We select the middle layers (23) of the model for editing. The forgetting boundaries \(b_u=S_u(\theta_0)-\Delta_{\mathrm f}\), with \(\Delta_{\mathrm f}=\log 2\). 
We use four NVIDIA RTX 4090 GPUs. Model-specific settings are provided in Appendix

\subsection{Main Results}
\vspace{-5pt}
\begin{wraptable}{r}{0.48\columnwidth}
\vspace{-20pt}
\centering
\caption{Copyright unlearning results.}
\resizebox{\linewidth}{!}{
\begin{tabular}{lcccc@{}}
\toprule
\multirow{2}{*}{\textbf{Method}}
& \multicolumn{2}{c}{Forget Perf.}
& \multicolumn{2}{c}{Retain Perf.} \\
\cmidrule(lr){2-3}
\cmidrule(lr){4-5}
& {BLEU$\downarrow$}
& {R-L$\downarrow$}
& {MMLU$\uparrow$}
& {Flu.$\uparrow$} \\
\midrule
Original & 74.76 & 98.68 & 46.39 & 4.03 \\
WHP & 23.55 & 17.93  & 38.49 & 2.52 \\
ICUL & 38.47 & 32.49 & 39.86 & \textbf{3.60} \\
SCANS & 19.92 & 17.72  & \underline{42.05} & 3.27 \\
ALTER & 12.96 & {10.40} & {41.84} & 3.08 \\
ASU & \underline{10.72} & \underline{9.68} & 37.56 &  3.28 \\
PTP-U  & \textbf{8.61} & \textbf{6.33}  & \textbf{43.34} & \underline{3.34} \\
\bottomrule
\end{tabular}
}
\label{tab:main_muse_books}
\vspace{-5pt}
\end{wraptable}
\textbf{Forget-retain trade-off.} Table~\ref{tab:main_muse_books} evaluates long-form copyright continuation on Llama2-7B. PTP-U achieves the lowest BLEU and R-L while preserving the highest MMLU, indicating a strong balance between forgetting and retained capability. Table~\ref{tab:main_rwku_wmdp} further evaluates this trade-off in two restricted-knowledge settings. RWKU measures entity forgetting with FB and QA, and AA evaluates robustness to adversarial recovery. WMDP measures Bio/Cyber forgetting and retained general capability. On Llama3.1-8B-Instruct, PTP-U preserves 65.54 Gen. on RWKU and 63.80 on WMDP. ALTER and ASU achieve stronger suppression on several direct forgetting metrics, but cause larger utility degradation and weaker AA robustness on RWKU. Representation intervention and parameter editing are generally more robust to adversarial recovery, but often reduce general capability substantially. ICUL preserves fluency through inference-time prompting but provides weaker forgetting. Overall, PTP-U maintains a favorable trade-off. Its local analytic proposal satisfies forgetting constraints, while nonlinear policy projection reduces drift from the Base model without relaxing them. This separation enables recovery from behavioral changes caused by the proposal across  unlearning tasks. Case analyses are in Appendix

\begin{table*}[t]
\centering
\caption{Main results on RWKU and WMDP. 
\textsuperscript{$\dagger$}, \textsuperscript{$\ddagger$}, and \textsuperscript{$\star$} 
denote output-distribution optimization, representation intervention \& parameter editing, and prompt-based unlearning methods, respectively.}
\begin{tabular}{@{}lccccc|ccc@{}}
\toprule
\multirow{2}{*}{\textbf{Baseline}}
& \multicolumn{5}{c}{\textbf{RWKU}}
& \multicolumn{3}{c}{\textbf{WMDP}} \\
\cmidrule(lr){2-6}\cmidrule(lr){7-9}
& \textbf{FB$\downarrow$}
& \textbf{QA$\downarrow$}
& \textbf{AA$\downarrow$}
& \textbf{Gen.$\uparrow$}
& \textbf{Flu.$\uparrow$}
& \textbf{Bio.$\downarrow$}
& \textbf{Cyber$\downarrow$}
& \textbf{Gen.$\uparrow$} \\
\midrule

\rowcolor{gray!20}
\multicolumn{9}{c}{\textit{\textbf{Llama3.1-8B-Instruct~\citep{grattafiori2024llama}}}} \\
\midrule
Base
& 63.95 & 66.42 & 69.83 & 68.37 & 4.20 & 72.74 & 47.35 & 68.37 \\
ICUL\textsuperscript{$\star$}
& 43.23 & 32.57 & 42.60 & \underline{64.80} & \textbf{4.07} & 45.90 & 33.34 & 61.10 \\
RMU\textsuperscript{$\ddagger$}
& 19.10 & 23.84 & 21.30 & 54.71 & 2.96 & 32.52 & 32.75 & 53.43 \\
MET\textsuperscript{$\ddagger$}
& 33.64 & 28.47 & 23.54 & 58.52 & 3.18 & 34.63 & 32.47 & 52.19 \\
SCANS\textsuperscript{$\ddagger$}
 & 16.80 & 23.47 & \underline{21.05} & 60.40 & 3.61 & 31.70 & 34.00 & 58.20 \\
\midrule
NPO$_{KL}$\textsuperscript{$\dagger$}
& 36.02 & 29.10 & 26.89 & 60.29 & 3.30 & 38.38 & 35.32 & 58.37 \\
ALTER\textsuperscript{$\dagger$}
& 18.61 & 22.51 & 22.95 & 61.38 & 3.20 & 29.69 & \underline{31.77} & \underline{61.60} \\
ASU\textsuperscript{$\dagger$}
& \textbf{8.90} & \textbf{10.62} & 26.80 & 60.89 & 3.43 & \textbf{27.45} & 32.48 & 57.74 \\
PTP-U\textsuperscript{$\dagger \ddagger$}
& \underline{11.42} & \underline{13.08} & \textbf{15.20} & \textbf{65.54} & \underline{3.93} & \underline{28.14} & \textbf{29.84} & \textbf{63.80} \\
\midrule
\rowcolor{gray!20}
\multicolumn{9}{c}{\textit{\textbf{Qwen3-14B~\citep{yang2025qwen3}}}} \\
\midrule
Base
& 63.06 & 45.42 & 56.02 & 79.35 & 4.41 & 75.51 & 61.06 & 79.35 \\
ICUL\textsuperscript{$\star$}
& 37.23 & 31.57 & 32.60 & 69.83 & \textbf{4.22} & 46.82 & 31.50 & 68.22 \\
RMU\textsuperscript{$\ddagger$}
& 22.15 & 25.71 & \textbf{14.33} & 62.30 & 2.96 & 43.85 & 36.24 & 66.51 \\
SCANS\textsuperscript{$\ddagger$}
& 21.50 & 28.11 & 26.40 & 68.70 & 3.79 & 34.60 & 35.80 & 67.50 \\
ALTER\textsuperscript{$\dagger$}
& 20.60 & 23.47 & 27.87 & \underline{71.02} & 3.35 & 34.24 & \textbf{30.58} & \underline{70.55} \\
ASU\textsuperscript{$\dagger$}
& \textbf{11.30} & \underline{15.10} & 24.94 & 69.53 & 3.43 & \textbf{29.40} & 31.58 & 68.38 \\
PTP-U\textsuperscript{$\dagger \ddagger$}
& \underline{13.76} & \textbf{14.55} & \underline{15.84} & \textbf{72.62} & \underline{3.90} & \underline{30.14} & \underline{31.23} & \textbf{72.42} \\
\bottomrule
\end{tabular}
\vspace{-5pt}
\label{tab:main_rwku_wmdp}
\end{table*}

\begin{figure}[h]
\centering
\includegraphics[width=0.99\linewidth]{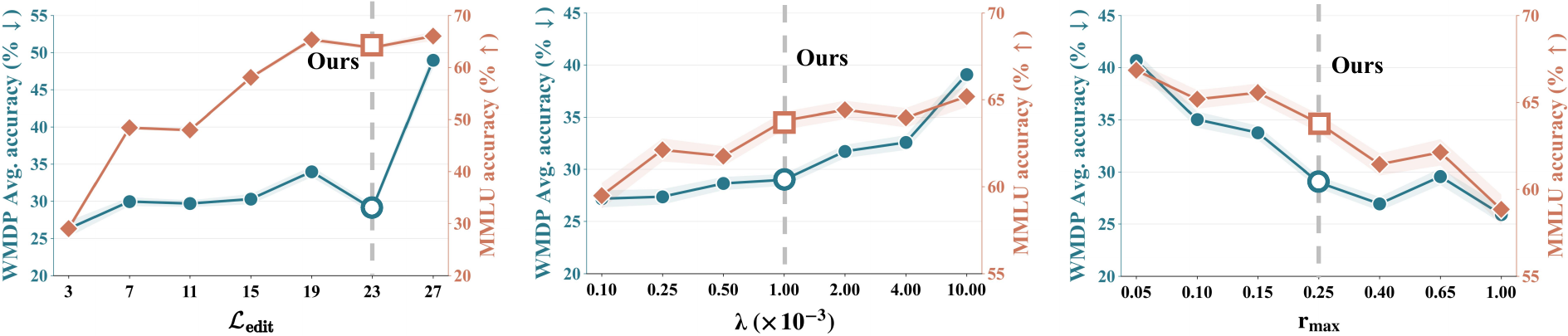}
\caption{Hyperparameter sensitivity on WMDP. Editing window $\mathcal{L}_{\mathrm{edit}}$, indexed by its last layer; each window contains three consecutive layers. Closer to 25\% indicates better forgetting, and orange diamonds show that higher MMLU accuracy is better.}
\label{Fig_Main:hyper}
\end{figure}

\textbf{Hyperparameter sensitivity.}
In figure~\ref{Fig_Main:hyper}, we examine $\mathcal{L}_{\mathrm{edit}}$, $\lambda$, and $r_{\max}$ by varying one factor while fixing the others at the reference configuration. The layer axis denotes the final layer of a fixed three-layer editing window. The favorable trade-off near layer 23 suggests that this middle-to-late window provides a more selective intervention point for knowledge-dependent computations, rather than a general advantage of deeper editing. The other hyperparameters control Stage~I. Specifically, $\lambda$ regularizes the local curvature metric, while $r_{\max}$ limits constraint-residual reduction per analytic iteration. Across the tested range, increasing $\lambda$ generally improves retention at the expense of forgetting. Increasing $r_{\max}$ from 0.05 to 1.00 reduces WMDP accuracy from 40.68\% to 25.92\%, closer to chance, while decreasing MMLU from 66.84\% to 58.84\%, indicating greater collateral damage under aggressive target advancement. As $r_{\max}\rightarrow0$, the required positive progress on violated constraints vanishes. As $\lambda\rightarrow0$, curvature damping disappears and numerical stability may degrade. Neither limit necessarily disables the full pipeline. We therefore use the window ending at layer 23, $\lambda=10^{-3}$, and $r_{\max}=0.25$, yielding 28.99\% WMDP and 63.80\% MMLU accuracy. Additional Stage~II analyses and internal visualizations are provided in Appendix

\textbf{Ablation of Control Components.}
In Table~\ref{tab:component_ablation}, omitting Stage~II reduces MMLU by $5.88$ points relative to full PTP-U, with little change in WMDP accuracy,, suggesting that the analytic edits leave recoverable capability loss.
Removing $\mathcal C_f$ from Stage~II yields the highest MMLU but increases WMDP accuracy from $28.99$ to $44.62$.
Thus, unconstrained recovery can improve retained performance while weakening forgetting.
Omitting Stage~I worsens both metrics, suggesting that analytic initialization facilitates subsequent recovery.
Setting $\alpha_t=1$ removes the current-policy proximity term and reduces retention despite preserving the forgetting constraints, indicating an additional contribution from current-policy regularization.
Under the same Stage~II protocol, Table~\ref{tab:Ablation_StageI} shows that fixed Attention$\rightarrow$FFN updates do not outperform joint editing, whereas refreshing the FFN local model improves both metrics.
The difference between these sequential variants supports recomputing the FFN update after the Attention edit changes
its inputs, linking the observed gains to relinearization within the chosen update order. The RWKU ablation shows in Appendix

\begin{table}[t]
\centering
\begin{minipage}[t]{0.48\textwidth}
\centering
\caption{Ablation of the two-stage framework. Removing \(\mathcal C_f\) applies only to Stage II. \(\alpha_t=1\) retains the forgetting constraints.}
\setlength{\tabcolsep}{2.5pt}
\begin{tabular}{lcc}
\toprule
\textbf{Method} & \textbf{WMDP Avg.$\downarrow$}
& \textbf{MMLU$\uparrow$} \\
\midrule
\textbf{PTP-U} & 28.99$\pm$0.79  & 63.80$\pm$0.68 \\
w/o Stage I & 34.31$\pm$1.42 & 60.26$\pm$1.21  \\
w/o Stage II & 29.23$\pm$1.18 & 57.92$\pm$1.31  \\
w/o \(\mathcal C_f\) & 44.62$\pm$1.56 & 64.84$\pm$0.57 \\
\(\alpha_t=1\) & 29.81$\pm$0.93 & 60.72$\pm$0.86 \\
\bottomrule
\end{tabular}
\label{tab:component_ablation}
\end{minipage}
\hfill
\begin{minipage}[t]{0.48\textwidth}
\centering
\caption{Ablation of Stage \textit{I} update ordering. All variants use the same subsequent Stage \textit{II} correction protocol.}
\setlength{\tabcolsep}{1.5pt}
\begin{tabular}{lcc}
\toprule
\textbf{Method} & \textbf{WMDP Avg.$\downarrow$}
& \textbf{MMLU$\uparrow$}  \\
\midrule
Attention-only  & 36.70$\pm$1.53 & 60.84$\pm$1.17  \\
FFN-only  & 33.36$\pm$1.36 & 59.12$\pm$0.94  \\
Joint & 31.04$\pm$1.08 & 61.81$\pm$0.83
\\
A $\rightarrow$ F fixed & 32.13$\pm$1.24 & 61.76$\pm$1.02 \\
\textbf{A $\rightarrow$ F refreshed} & 28.99$\pm$0.79 & 63.80$\pm$0.68 \\
\bottomrule
\end{tabular}
\label{tab:Ablation_StageI}
\end{minipage}
\vspace{-10pt}
\end{table}

\section{Discussions}

\begin{wrapfigure}{r}{0.45\textwidth}
\centering
\includegraphics[width=1.0\linewidth]{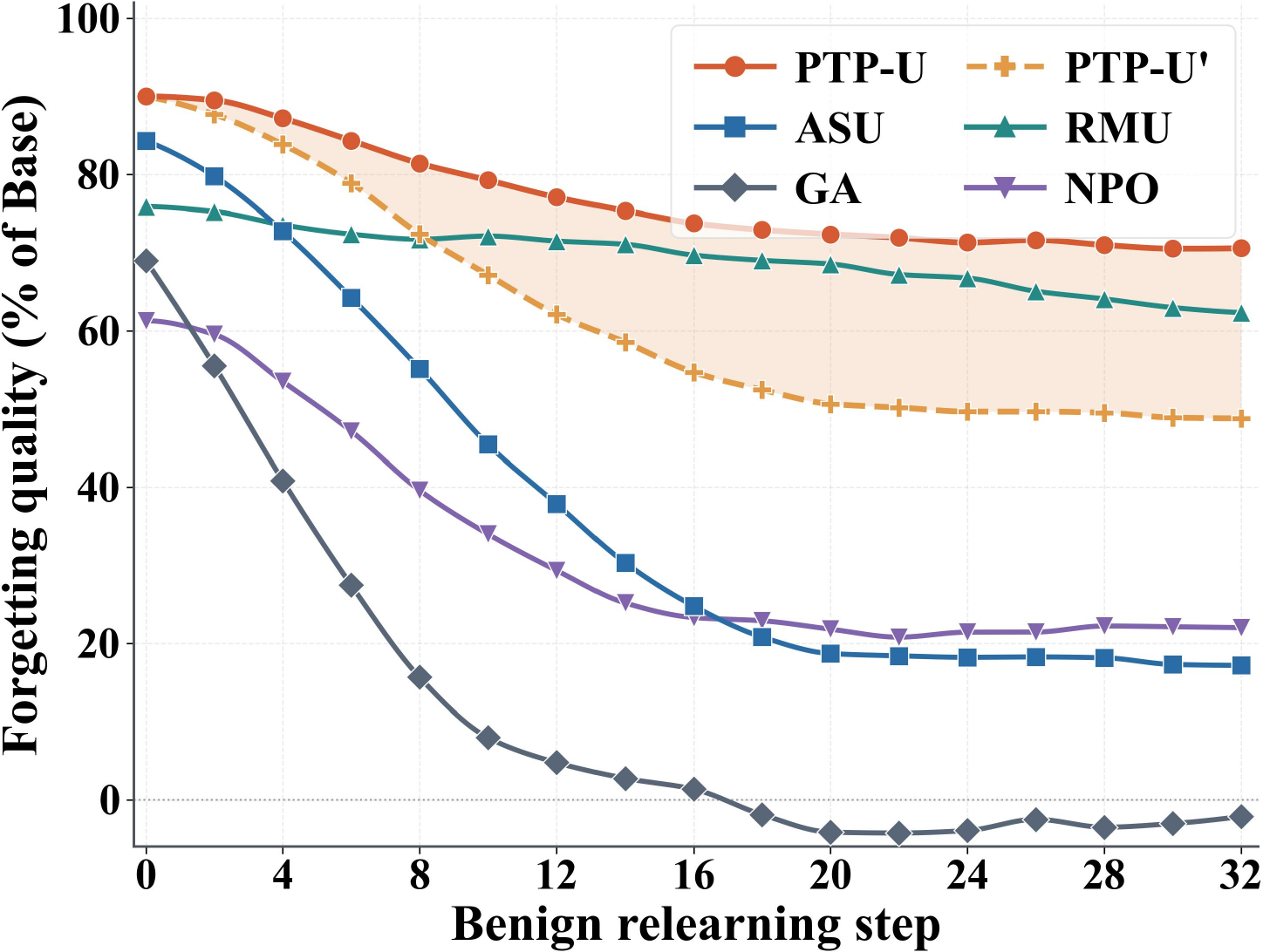}
\caption{Post-unlearning benign relearning attack. Shading highlights the gap between the PTP-U and PTP-U' variant.}
\label{fig:benign_relearning}
\vspace{-10pt} 
\end{wrapfigure}
\textbf{Why Forgetting Constraints Matter for Recovery.} Stage II moves toward the Base policy to restore non-target capability while preserving forgetting. Figure~\ref{Fig_Main:Obs} shows that non-target capability can improve with limited degradation in forgetting quality, raising whether an explicit forgetting constraint is necessary. Benign relearning~\citep{hu2025unlearning} shows that benign fine-tuning may revive suppressed knowledge. Figure~\ref{fig:benign_relearning} shows that after 32 steps, PTP-U retains 70.23\% forgetting quality, compared with 48.42\% for PTP-U' without the forgetting constraint under the same KL regularization. Explicit forgetting control therefore limits knowledge revival. ASU drops to 17.17\% without an additional forgetting guard, suggesting that attention-based smoothing remains vulnerable to parameter updates. RMU retains 62.33\%, indicating more persistent representational changes, but with lower general capability. PTP-U instead balances capability restoration and forgetting protection. Table~\ref{tab:main_rwku_wmdp} further reports RWKU AA performance across different models together with general capability. Benign relearning and AA provide complementary views of PTP-U's robustness against knowledge revival from parameter updates and adversarial prompting. Relearning settings are provided in Appendix

\textbf{Preservation of Neighborhood Knowledge.}
We evaluate whether the modifications introduced by unlearning propagate to intact knowledge that ought to be preserved.
We partition the TOFU evaluation~\citep{maini2024tofu} into target unlearning data (TUD), neighborhood knowledge from the same domain (NEK), and general external knowledge (GEK), corresponding to Forget10, Real Authors, and World Facts, respectively. This separation distinguishes target forgetting from local interference and broader capability degradation. As shown in Table~\ref{tab:maintofu_performance}, the prompting-based ICUL achieves the highest NEK score of 75.0\%, reflecting the preservation benefit of avoiding parameter modification. However, its TUD R-L remains 0.27, indicating still reproduces the target facts to be forgotten. Compared with RMU, PTP-U reduces target-content reproduction while improving both NEK and GEK by 15.0 percentage points. These results support selective unlearning, where the two stages reduce target-knowledge recoverability while limiting interference with shared knowledge. Complete experimental settings and analyses are provided in Appendix

\textbf{Cost analysis.}
Table~\ref{tab:cost_analysis} illustrates the offline and online overheads. 
Memory denotes peak aggregate allocation across active devices, whereas serving latency refers to an individual query on a single-GPU replica. PTP-U requires 31.8 minutes for model editing and 3.7 minutes for nonlinear fine-tuning.
Its principal offline overhead comes from geometry construction, sequential relinearization, and actual-model verification, rather than the reduced linear solve itself.
The 42.7 GB memory requirement reflects these additional statistics and intermediate tensors; methods such as RMU also incur substantial memory costs when maintaining both an updated model and a frozen reference.
PTP-U does not minimize offline adaptation cost.
Instead, its parameter updates can be merged into the backbone, eliminating auxiliary inference modules and yielding a serving latency comparable to other weight-updating methods.
ICUL avoids offline adaptation but prepends additional examples to each query, increasing prefill computation and KV-cache usage, with further decoding overhead from the longer context.
Its serving cost therefore depends on demonstration length and prefix-cache reuse.
Instead, PTP-U trades additional offline computation for a standard deployment interface without recurring prompt expansion.
\vspace{-10pt}
\begin{table}[h]
\centering
\begin{minipage}[t]{0.40\textwidth}
\centering
\caption{TOFU (10\%) with Llama-3.1-8B-Instruct.}
\setlength{\tabcolsep}{6.5pt}
\begin{tabular}{@{}lccc@{}}
\toprule
\multirow{2}{*}{\textbf{Method}}
& \textbf{TUD}
& \textbf{NEK}
& \textbf{GEK} \\
& \textbf{R-L$\downarrow$}
& \textbf{Acc $\uparrow$}
& \textbf{Acc $\uparrow$} \\
\midrule
NPO\_KL & 0.31 & 0.63 & 0.68 \\
ICUL    & 0.27 & \textbf{0.75} & \underline{0.74} \\
RMU     & 0.19 & 0.59 & 0.64 \\
MET     & 0.18 & 0.63 & 0.66 \\
ALTER   & 0.15 & 0.71 & 0.72 \\
ASU     & \underline{0.14} & 0.69 & 0.71 \\
\midrule
PTP-U   & \textbf{0.10} & \underline{0.74} & \textbf{0.79} \\
\bottomrule
\end{tabular}
\label{tab:maintofu_performance}
\end{minipage}
\hfill
\begin{minipage}[t]{0.58\textwidth}
\centering
\caption{Model-side adaptation and serving costs on Llama-3.1-8B-Instruct.
Memory denotes peak GB.
Latency measures free-form generation, not multiple-choice scoring.
``-'' indicates no offline adaptation.}
\setlength{\tabcolsep}{1.5pt}
\begin{tabular}{lccc}
\toprule
\textbf{Method}
& \textbf{Time (min)}
& \textbf{Memory (GB)}
& \textbf{Latency (s/query)} \\
\midrule
NPO\_KL & 17.3       & 23.1 & 1.20 \\
RMU     & 24.9       & 38.0 & 1.20 \\
ASU      & 15.3       & 20.4 & 1.21 \\
ALTER   & 9.2        & 18.3 & 1.23 \\
ICUL    & -          & -    & 1.86 \\
PTP-U   & 35.5       & 42.7 & 1.20 \\
\bottomrule
\end{tabular}
\label{tab:cost_analysis}
\end{minipage}
\vspace{-10pt}
\end{table}

\vspace{-5pt}
\section{Related Work}
\vspace{-5pt}

\textbf{LLM unlearning.}
LLM unlearning methods broadly operate through output optimization, internal intervention, or inference-time prompting. Output-based methods modify target likelihoods or preferences~\citep{fan2025simplicity,xu2025relearn,li2026cipo}, while representation- and editing-based methods localize changes to internal features or parameter subsets~\citep{shi2025safety,li2026editing}. Prompt-based approaches instead suppress target behavior through test-time context without permanently modifying model parameters~\citep{pawelczyk2024context,takashiro2025answer,wang2026dragon}. Recent work has further explored more controlled model modification through Fisher-aware parameter selection~\citep{cha2025towards}, minimum-norm editing~\citep{guo2025simulating}, and explicit optimization constraints~\citep{dhillon2025l3ms,entesari2025constrained}. Across these approaches, a recurring challenge is achieving the desired target modification while preserving non-target behavior.

\textbf{Post-training preservation and locality.}
Recent studies have examined how post-training alters pretrained capabilities beyond the optimized task. Smaller policy shifts from the Base model have been associated with better preservation of prior capabilities~\citep{shenfeld2026rl}, while broader analyses reveal heterogeneous forgetting across examples and tasks~\citep{harmon2026mapping}. In knowledge removal, mechanistic localization can reduce unintended side effects~\citep{guo2025mechanistic}, although parameter locality alone does not necessarily imply effective or isolated unlearning~\citep{lee2025localization}. These findings motivate evaluating model modification not only by whether the target behavior changes, but also by how broadly the update affects the original model.

In general, existing work shows that successful unlearning does not uniquely determine how a model should change. Different optimization paths may achieve similar forgetting while inducing different degrees of non-target drift. This motivates examining not only whether forgetting is achieved, but also which induced changes are actually necessary.

\vspace{-5pt}
\section{Conclusion and Limitation}
\vspace{-5pt}

We presented PTP-U, which combines curvature-guided analytic editing with nonlinear policy recovery under fixed forgetting constraints. Sequential Attention-to-FFN relinearization updates downstream approximations, while Base-referenced distillation reduces collateral behavioral drift. Experiments across four benchmarks support the recoverable-drift premise: some unlearning-induced changes can be reversed without substantially weakening observed forgetting. More broadly, PTP-U connects model editing and teacher-guided distillation through a common constrained objective, using local geometry to propose edits and policy-space proximity to guide recovery. This perspective highlights the value of geometric optimization for distinguishing necessary target modification from avoidable capability degradation. However, local curvature approximations and constraints defined over sampled knowledge units do not guarantee globally minimal behavioral drift or the elimination of unobserved knowledge-retrieval routes. Future research will investigate tighter nonlinear approximations and semantic constraints that generalize beyond observed queries. We will also study their applicability to multimodal semantic unlearning, where cross-modal knowledge sharing introduces additional challenges in separating target associations from representations essential to retained capabilities.

\bibliography{iclr2027_conference}
\bibliographystyle{iclr2027_conference}

\end{document}